\documentclass[10pt,twocolumn,letterpaper]{article}

\usepackage{iccv}
\usepackage{times}
\usepackage{epsfig}
\usepackage{graphicx}
\usepackage{amsmath}
\usepackage{amssymb}

\usepackage{cite}
\usepackage{amsmath,graphicx}
\usepackage{epstopdf}
\usepackage{xcolor}
\usepackage{amssymb}
\usepackage{array}
\usepackage{booktabs}
\usepackage{multirow}
\usepackage{graphicx}
\usepackage{color}
\usepackage{float}
\usepackage{stfloats}
\usepackage[misc]{ifsym}
\usepackage{stmaryrd}
\usepackage{grffile}
\usepackage{subfig}
\usepackage{makecell}
\usepackage{float}
\usepackage{threeparttable}
\usepackage{enumitem}

\usepackage[pagebackref=true,breaklinks=true,letterpaper=true,colorlinks,bookmarks=false]{hyperref}

\iccvfinalcopy 

\ificcvfinal\fi

\begin{document}

\title{Flickr1024: A Large-Scale Dataset for Stereo Image Super-Resolution}

\author{Yingqian Wang\textsuperscript{1}, Longguang Wang\textsuperscript{1}, Jungang Yang\textsuperscript{1*}, Wei An\textsuperscript{1}, and Yulan Guo\textsuperscript{1}\\
\textsuperscript{1}College of Electronic Science and Technology, National University of Defense Technology, China\\
{\tt\small \{wangyingqian16, yangjungang\}@nudt.edu.cn}
}

\maketitle
\ificcvfinal\thispagestyle{empty}\fi

\begin{abstract}
With the popularity of dual cameras in recently released smart phones, a growing number of super-resolution (SR) methods have been proposed to enhance the resolution of stereo image pairs. However, the lack of high-quality stereo datasets has limited the research in this area. To facilitate the training and evaluation of novel stereo SR algorithms, in this paper, we present a large-scale stereo dataset named \textit{Flickr1024}, which contains 1024 pairs of high-quality images and covers diverse scenarios. We first introduce the data acquisition and processing pipeline, and then compare several popular stereo datasets. Finally, we conduct cross-dataset experiments to investigate the potential benefits introduced by our dataset. Experimental results show that, as compared to the \textit{KITTI} and \textit{Middlebury} datasets, our \textit{Flickr1024} dataset can help to handle the over-fitting problem and significantly improves the performance of stereo SR methods. The \textit{Flickr1024} dataset is available online at: https://yingqianwang.github.io/Flickr1024.
\end{abstract}

\section{Introduction}\label{sec1}

With recent advances in camera miniaturization, dual cameras are commonly adopted in commercial mobile phones. Using the complementary information provided by binocular systems, the resolution of image pairs can be enhanced by stereo super-resolution (SR) methods \cite{bhavsar2010resolution,jeon2018enhancing,wang2019learning}. Nowadays, many top-performing SR methods \cite{jeon2018enhancing,wang2019learning,wang2018learning,zhang2019residual} are built upon deep neural networks. These data-driven SR methods can be enormously benefited from large-scale high-quality datasets such as \textit{DIV2K}\cite{agustsson2017ntire} and \textit{Vimeo-90K}\cite{xue2017video}.

In the area of stereo vision, several datasets are currently available. The \textit{KITTI} stereo datasets\cite{Geiger2012CVPR,menze2015object} are mainly developed for autonomous driving. All images in the \textit{KITTI2012}\cite{Geiger2012CVPR} and \textit{KITTI2015}\cite{menze2015object} datasets are captured by two video cameras mounted on the top of a car. The scenes in the \textit{KITTI} datasets only include roads or highways from driving perspectives. Groundtruth disparity is provided for the training of stereo matching and visual odometry.
The \textit{Middlebury} stereo dataset consists of a series of sub-datasets introduced in 2003\cite{scharstein2003high}, 2005\cite{scharstein2007learning}, 2006\cite{hirschmuller2007evaluation}, and 2014\cite{scharstein2014high}. The \textit{Middlebury} dataset is acquired in the laboratory, its scenes only cover close-shots of different objects. Note that, 55 of its 65 image pairs are provided with groundtruth disparity for stereo matching.
The \textit{ETH3D} stereo dataset is a part of the \textit{ETH3D} benchmark\cite{schops2017multi}. Groundtruth depth is provided for visual odometry and 3D reconstruction. Note that, images in the \textit{ETH3D} dataset are of gray scale, of low resolution, and with limited scenarios.

\begin{figure}[t]
\vspace{0.9cm}
\centering
\includegraphics[width=8cm]{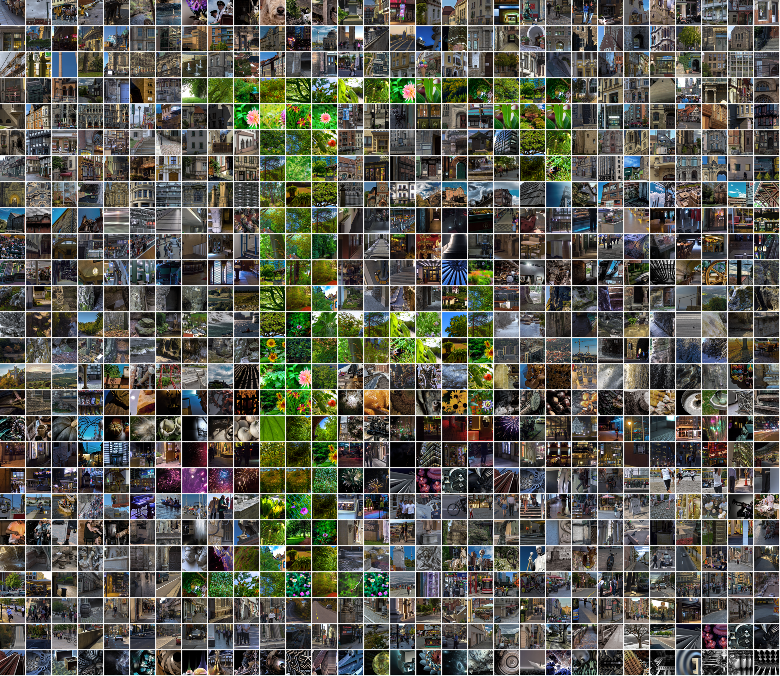}
\caption{{The \textit{Flickr1024} dataset.}\label{fig1}}
\vspace{0.2cm}
\end{figure}

\begin{figure*}[t]
\centering
\includegraphics[width=16cm]{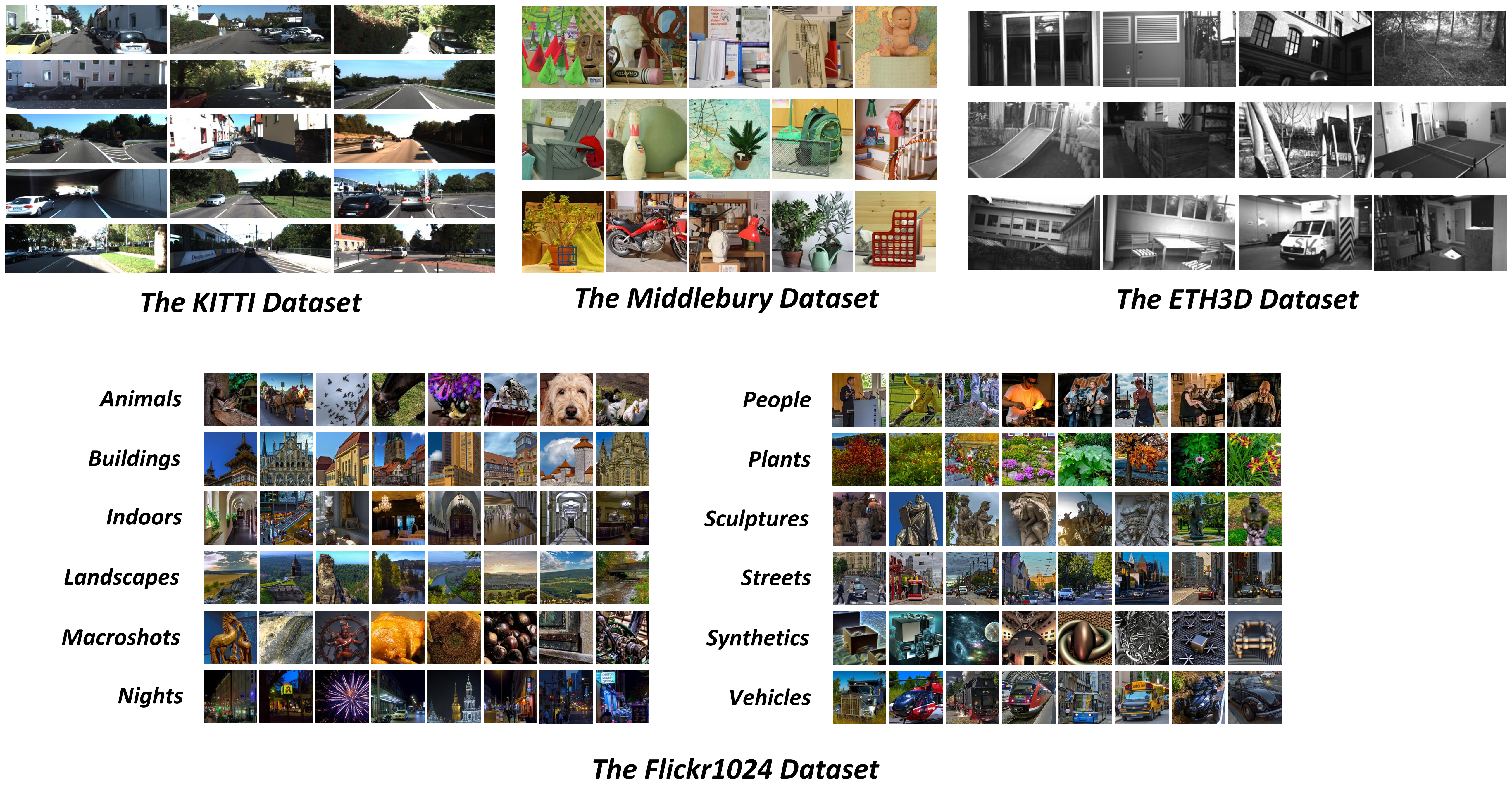}
\vspace{0.3cm}
\caption{{Example images sampled from several popular stereo datasets: \textit{KITTI}\cite{Geiger2012CVPR,menze2015object}, \textit{Middlebury}\cite{scharstein2003high,scharstein2007learning,hirschmuller2007evaluation,scharstein2014high}, \textit{ETH3D}\cite{schops2017multi}, and \textit{Flickr1024}. It can be observed that our \textit{Flickr1024} dataset covers significantly more diverse scenarios as compared to the existing stereo datasets.}\label{fig2}}
\end{figure*}

Since the task of stereo vision can vary significantly (e.g., stereo matching \cite{liang2018learning}, stereo segmentation \cite{li2018learning}), existing stereo datasets are unsuitable for stereo SR due to the insufficient number of images and limited types of scenarios. To design, train, and evaluate novel stereo SR methods, a large-scale and high-quality stereo dataset with diverse scenarios is highly needed. In this paper, we present a new \textit{Flickr1024} dataset (see Fig.~\ref{fig1}) for stereo SR. In summary, the contributions of our dataset can be listed as follows:

\begin{itemize}
\item It is the largest dataset for stereo SR to date, which contains 1024 high-quality image pairs and covers diverse scenarios.
\item The scenarios covered by our dataset are highly consistent with real cases in daily photography (see Fig.~\ref{fig2}). Consequently, algorithms developed on the \textit{Flickr1024} dataset can be easily adopted in real-world applications such as mobile phones.
\item Experimental results demonstrate that our dataset can help to address the over-fitting problem and significantly improve the performance of different stereo SR methods. That is, our dataset can benefit both industrial and research communities in stereo SR.
\end{itemize}

The \textit{Flickr1024} dataset was initially introduced in our previous conference paper \cite{wang2019learning} and used as the augmented training data for our \textit{PASSRnet} algorithm. However, in our preliminary work, both details and effectiveness of this dataset have not been investigated. In this paper, we deeply investigate the \textit{Flickr1024} dataset and make several additional contributions, which can be summarized as follows:

\begin{itemize}
\item We provide more details in data acquisition and processing. The pipelines and processing methods used in this paper can help the community to develop new datasets for their own research.
\item We comprehensively compare the \textit{Flickr1024} dataset to several existing stereo datasets \cite{Geiger2012CVPR,menze2015object,schops2017multi,scharstein2003high,scharstein2007learning,hirschmuller2007evaluation,scharstein2014high} using different objective metrics. Evaluation results have demonstrated the superiority of our dataset.
\item We conduct cross-dataset experiment to study the performance gains introduced by the \textit{Flickr1024} dataset. Experimental results have demonstrated the effectiveness of our dataset in performance promotion and over-fitting elimination.
\end{itemize}

\section{Data Acquisition and Processing}\label{sec2}

To generate the \textit{Flickr1024} dataset, we manually collected 1024 RGB stereo photographs from albums on \textit{Flickr}$\footnote{https://www.flickr.com/}$ with the permissions of photograph owners. Since all images collected from \textit{Flickr} are in cross-eye pattern for 3D visualization, their optical axes should be corrected to be parallel. As shown in Fig.~\ref{fig3}, the processing pipeline can be summarized as follows:

\begin{figure}[t]
\vspace{0cm}
\centering
\includegraphics[width=8.5cm]{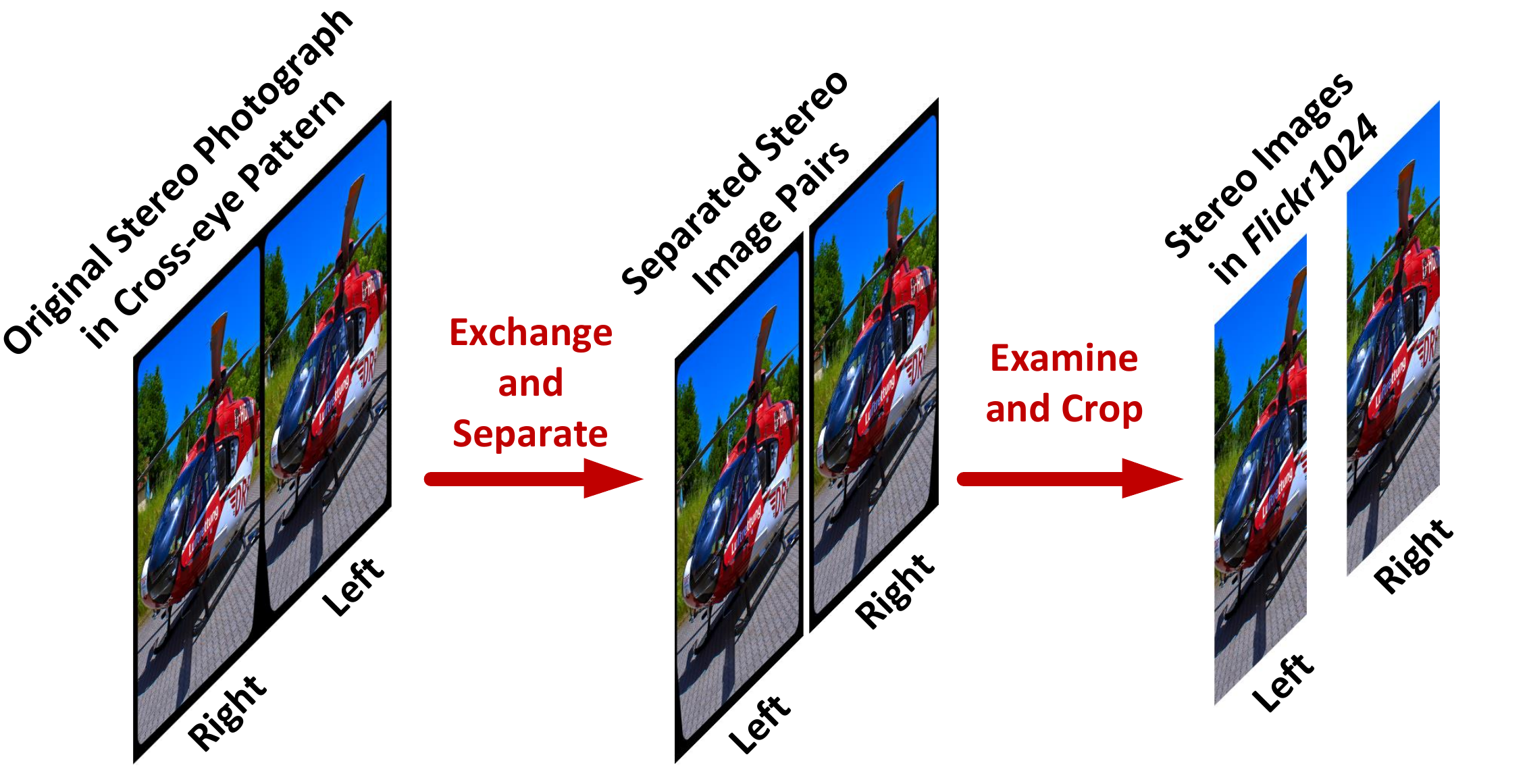}
\caption{{The processing pipeline to generate the \textit{Flickr1024} dataset.}\label{fig3}}
\end{figure}

\begin{table*}[t]
\renewcommand\arraystretch{1.3}
\centering
\footnotesize
\caption{Main characteristics of several popular stereo datasets. Both average value and standard deviation are reported. Among all the compared datasets, the \textit{Flickr1024} dataset achieves promising scores in image pairs, resolution, and perceptual image quality.}\label{tab1}
\vspace{-0cm}

\begin{tabular}{|>{}m{2.7cm}|>{\centering}m{1.4cm}>{\centering}m{2.4cm}>{\centering}m{1.6cm}
>{\centering}m{2.0cm}>{\centering}m{2.2cm}>{\centering}m{2.0cm}|}
\hline

Datasets & Image Pairs & Resolution ($\uparrow$)& Entropy ($\uparrow$)& BRISQE ($\downarrow$) \cite{mittal2012no}& SR-metric ($\uparrow$) \cite{ma2017learning}& ENIQA ($\downarrow$)\cite{chen2018noreference}\tabularnewline
\hline

\textit{KITTI2012}\cite{Geiger2012CVPR} & 389 & 0.46 ($\pm$0.00) Mpx & 7.12 ($\pm$0.30) & \textbf{17.49} ($\pm$6.56) & \textbf{7.15} ($\pm$0.63)&\underline{0.097} ($\pm$0.028)\tabularnewline

\textit{KITTI2015}\cite{menze2015object} & \underline{400} & 0.47 ($\pm$0.00) Mpx & 7.06 ($\pm$0.00) & 23.79 ($\pm$5.81) & 7.06 ($\pm$0.51)&0.169 ($\pm$0.030)\tabularnewline

\textit{Middlebury}\cite{scharstein2003high,scharstein2007learning,hirschmuller2007evaluation,scharstein2014high} & 65 & \textbf{3.59} ($\pm$2.06) Mpx & \textbf{7.55} ($\pm$0.20) & 26.85 ($\pm$13.30) & 6.01 ($\pm$1.08)& 0.270 ($\pm$0.120)\tabularnewline

\textit{ETH3D}\cite{schops2017multi} & 47 & 0.38 ($\pm$0.08) Mpx & \underline{7.24} ($\pm$0.43) & 27.95 ($\pm$12.06) & 5.99 ($\pm$1.52)&0.195 ($\pm$0.073)\tabularnewline

\textit{Flickr1024} & \textbf{1024} & \underline{0.73} ($\pm$0.33) Mpx & 7.23 ($\pm$0.64) & \underline{19.40} ($\pm$13.77) & \underline{7.12} ($\pm$0.67)&\textbf{0.065} ($\pm$0.073)\tabularnewline

\textit{Flickr1024} (Train) & 800 & 0.74 ($\pm$0.34) Mpx & 7.23 ($\pm$0.65) & 19.10 ($\pm$13.69) & 7.12 ($\pm$0.66) &0.063 ($\pm$0.074)\tabularnewline

\textit{Flickr1024} (Validation) & 112 & 0.72 ($\pm$0.23) Mpx & 7.26 ($\pm$0.54)& 20.03 ($\pm$12.54) &7.13 ($\pm$0.70)&0.074 ($\pm$0.084)\tabularnewline

\textit{Flickr1024} (Test) & 112 & 0.72 ($\pm$0.32) Mpx & 7.22 ($\pm$0.60) & 20.97 ($\pm$15.40) &7.12 ($\pm$0.67)&0.076 ($\pm$0.087)\tabularnewline
\hline

\end{tabular}
\begin{tablenotes}
    \footnotesize
    \item[1] Note: Mpx denotes megapixels per image. The best scores are in \textbf{bold} and the second best scores are \underline{underlined}.
\end{tablenotes}
\vspace{0.2cm}
\end{table*}

\begin{enumerate}
\item We cut each cross-eye photograph into a stereo image pair. Note that, to transform a cross-eye photograph into an image pair with parallel optical axis, the left and right images in the stereo image pair need to be exchanged.
\item We check each pair of stereo images to ensure that they are vertically rectified (i.e., image pairs has horizontal disparities only). In practice, most image pairs have already been calibrated in vertical direction by the photo owners. For these images without vertical calibration, we simply discard them from our dataset.
\item We crop the left and right images to remove black (or white) margins and to make zero disparity corresponding to infinite depth. Note that, regions with infinite depth are unavailable for close-shot images. We therefore, crop these image pairs to ensure that the minimum disparity is larger than a certain value (set to 40 pixels in our dataset).
\end{enumerate}

Finally, we randomly split our dataset to generate 800 training image pairs, 112 validation image pairs, and 112 test image pairs.

\section{Comparisons to Existing Datasets}\label{sec3}

\begin{table*}
\renewcommand\arraystretch{1.2}
\centering
\footnotesize
\caption{PSNR and SSIM values achieved by \textit{StereoSR}\cite{jeon2018enhancing} for 4$\times$~SR with 60 training epochs.}\label{tab2}
\vspace{-0cm}

\begin{tabular}{|>{\centering}m{3.2cm}|>{\centering}m{2.8cm}|>{\centering}m{2.8cm}|>{\centering}m{2.8cm}|>{\centering}m{2.8cm}|}
\hline

Dataset & \textit{KITTI2015} (Test) & \textit{Middlebury} (Test)& \textit{Flickr1024} (Test)& \textit{ETH3D} (Test)\tabularnewline
\hline

\textit{KITTI2015} (Train)& 24.28 / 0.741 & 26.27 / 0.749 & 21.77 / 0.617 & 29.63 / 0.831 \tabularnewline
\hline

\textit{Middlebury} (Train)& 23.64 / 0.743 & 26.62 / 0.773 & 21.64 / 0.646 & 28.66 / 0.843 \tabularnewline
\hline

\textit{Flickr1024} (Train)& \textbf{25.08} / \textbf{0.779} & \textbf{27.85} / \textbf{0.807} & \textbf{22.64} / \textbf{0.692} & \textbf{30.55} / \textbf{0.860} \tabularnewline
\hline
\end{tabular}
\vspace{0.0cm}
\end{table*}

\begin{table*}
\renewcommand\arraystretch{1.2}
\centering
\footnotesize
\caption{PSNR and SSIM values achieved by \textit{PASSRnet}\cite{wang2019learning} for 4$\times$~SR with 80 training epochs.}\label{tab3}
\vspace{-0cm}

\begin{tabular}{|>{\centering}m{3.2cm}|>{\centering}m{2.8cm}|>{\centering}m{2.8cm}|>{\centering}m{2.8cm}|>{\centering}m{2.8cm}|}
\hline

Dataset & \textit{KITTI2015} (Test) & \textit{Middlebury} (Test)& \textit{Flickr1024} (Test)& \textit{ETH3D} (Test)\tabularnewline
\hline

\textit{KITTI2015} (Train)& 23.13 / 0.703 & 25.42 / 0.762 & 21.31 / 0.600 & 26.95 / 0.789 \tabularnewline
\hline

\textit{Middlebury} (Train)& 25.18 / 0.774 & 28.08 / 0.853 & 22.54 / 0.676 & 31.39 / 0.864 \tabularnewline
\hline

\textit{Flickr1024} (Train)& \textbf{25.62} / \textbf{0.791} & \textbf{28.69} / \textbf{0.873} & \textbf{23.25} / \textbf{0.718} & \textbf{31.94} / \textbf{0.877} \tabularnewline
\hline

\end{tabular}
\vspace{-0.1cm}
\end{table*}

In this section, statistical comparisons are performed to demonstrate the superiority of the \textit{Flickr1024} dataset. The main characteristics of the \textit{Flickr1024} dataset and four existing stereo datasets \cite{Geiger2012CVPR,menze2015object,schops2017multi,scharstein2003high,scharstein2007learning,hirschmuller2007evaluation,scharstein2014high} are listed in Table~\ref{tab1}. Following \cite{agustsson2017ntire}, we use \textit{entropy} to measure the amount of information included in each dataset, and use three no-reference image quality assessment (NRIQA) metrics to assess the perceptual image quality, including blind/referenceless image spatial quality evaluator (BRISQE)\cite{mittal2012no}, SR-metric\cite{ma2017learning}, and entropy-based image quality assessment (ENIQA)\cite{chen2018noreference}. For image quality assessment, these NRIQA metrics are superior to many full-referenced measures (e.g., PSNR, RMSE, and SSIM), and highly correlated to human perception\cite{ma2017learning}. For all the NRIQA metrics presented in this paper, we run the codes provided by their authors under their original models and default settings. Note that, small values of BRISQE\cite{mittal2012no} and ENIQA\cite{chen2018noreference}, and large values of SR-metric\cite{ma2017learning} represent high image quality.

As listed in Table~\ref{tab1}, the \textit{Flickr1024} dataset is larger than other datasets by at least 2.5 times. Besides, the image resolution of the \textit{Flickr1024} dataset also outperforms that of the \textit{KITTI2012}, \textit{KITTI2015}, and \textit{ETH3D} datasets. Although the \textit{Middlebury} dataset has the highest image resolution, the number of image pairs in this dataset is limited. The entropy values of all datasets are comparable, while the entropy of the \textit{KITTI} datasets is relatively low. That is, the diversity of images in the \textit{KITTI} datasets is smaller than that of other datasets. For perceptual image quality assessment, both the \textit{Flickr1024} and the \textit{KITTI2012} datasets achieve promising scores. Specifically, the \textit{Flickr1024} dataset has the highest ENIQA score, and the second highest BRISQE and SRmetric scores. Since these metrics are influenced by the brightness and textures of tested images, the \textit{Flickr1024} dataset has higher standard deviations than existing datasets due to its diverse scenarios. These assessments indicate that images in \textit{Flickr1024} are relatively high in perceptual quality and suitable for stereo SR.

It is also notable that, comparable scores of these metrics can be achieved on three different subsets (i.e., training set, validation set, and test set) of the \textit{Flickr1024} dataset, as shown in Table~\ref{tab1}. That means, a good balance is achieved with random partition, and the bias between the training and the test process is relatively small.

\section{Cross-Dataset Evaluation}\label{sec4}

 To investigate the potential benefits of a large-scale dataset to the performance improvement of learning-based stereo SR methods, experimental results are provided in this section. Besides, a cross-dataset evaluation is performed to fully demonstrate the superiority of the \textit{Flickr1024} dataset.

\begin{figure*}[t]
\vspace{-0cm}
\centering
\subfloat[\textit{KITTI2015}]{\includegraphics[height=8cm]{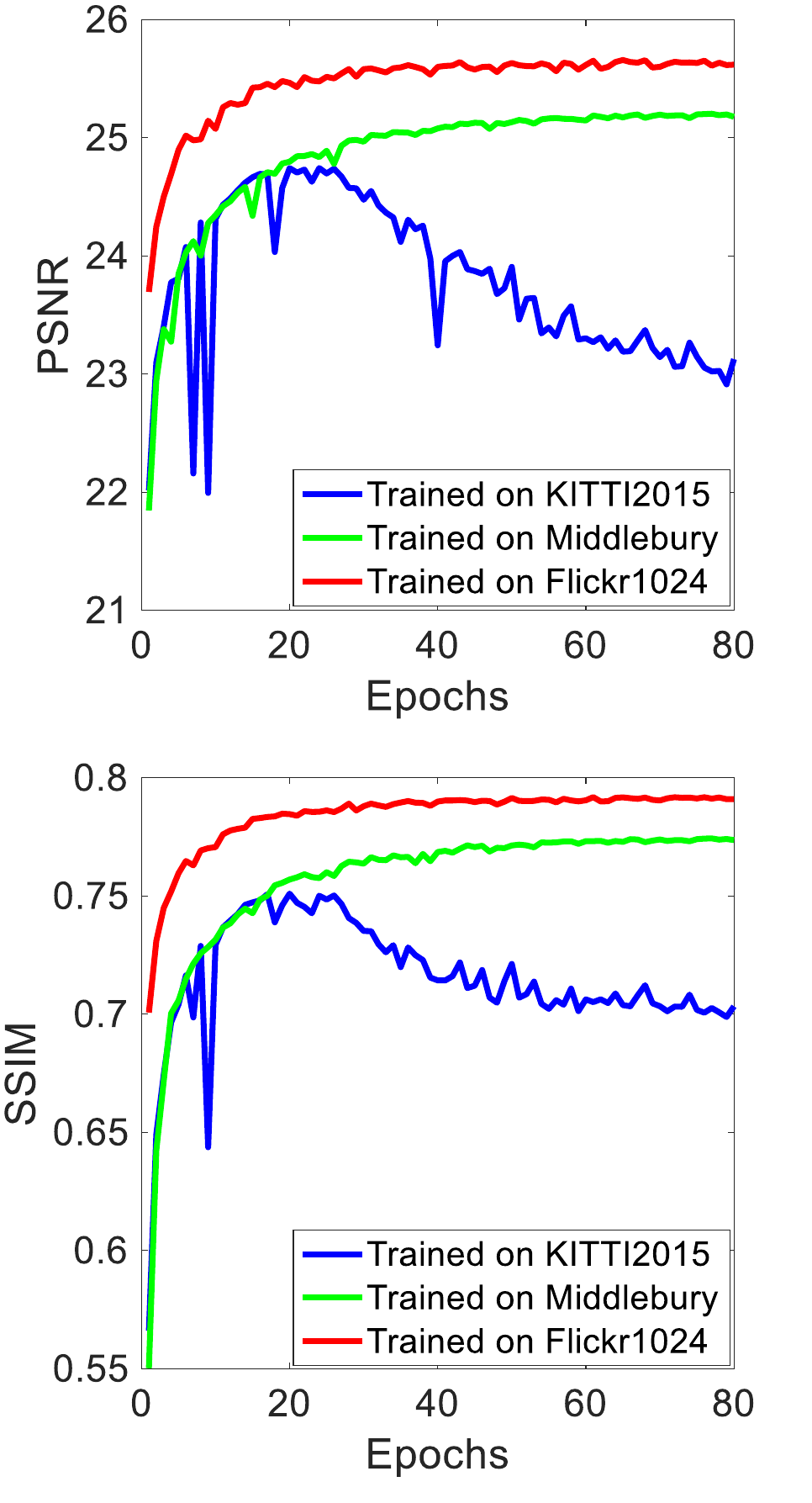}}
\subfloat[\textit{Middlebury}]{\includegraphics[height=8cm]{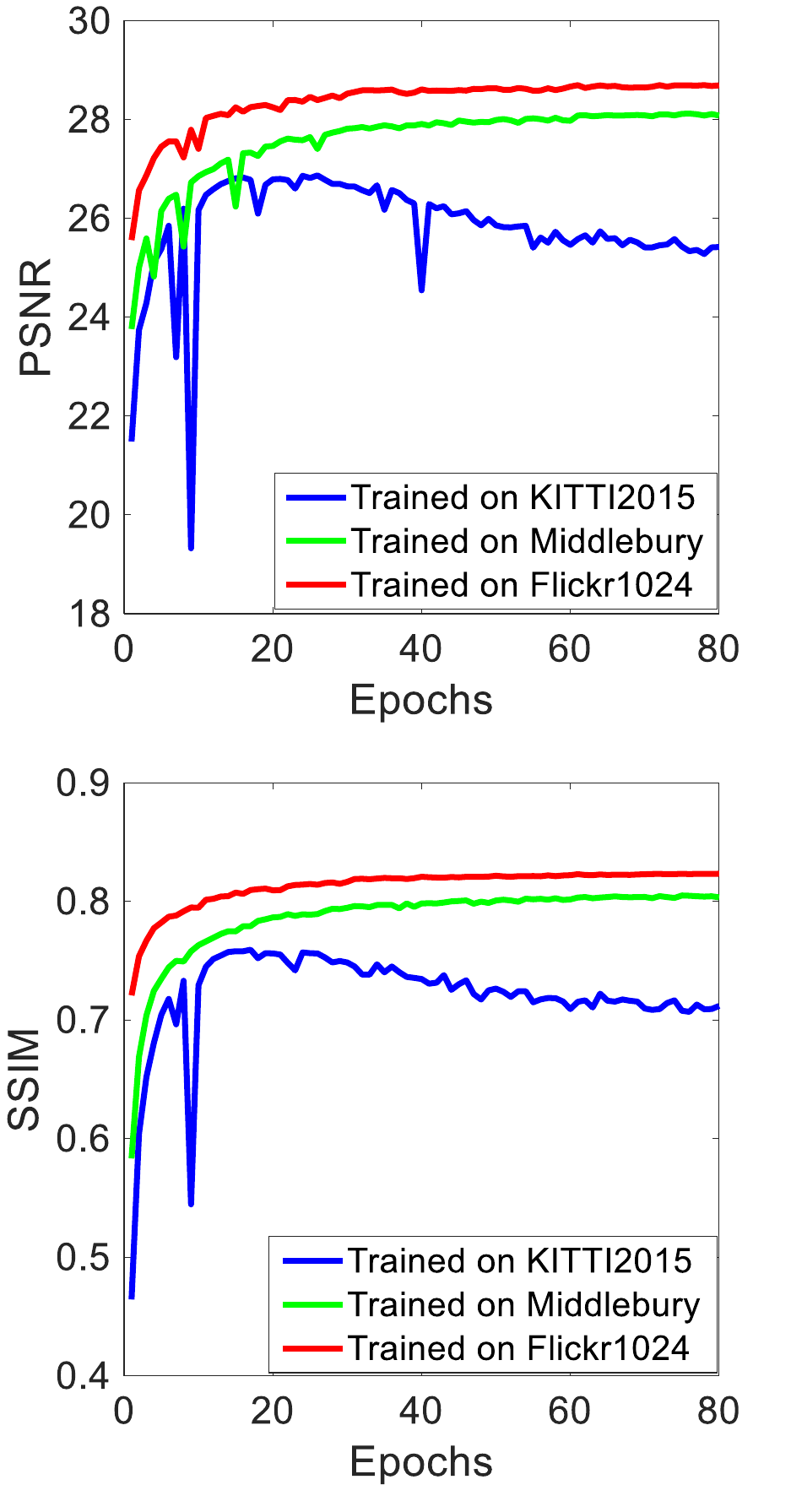}}
\subfloat[\textit{Flickr1024}]{\includegraphics[height=8cm]{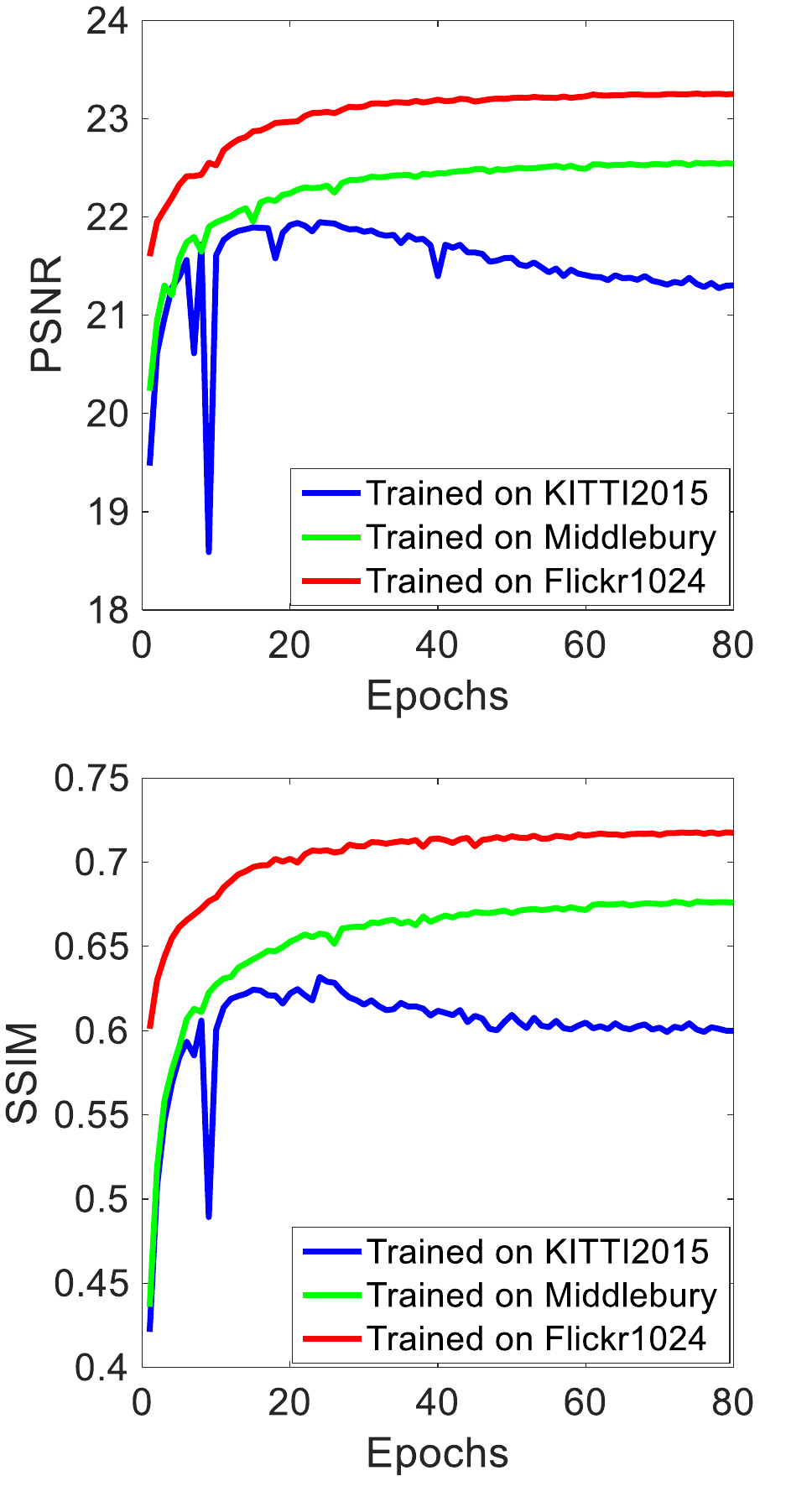}}
\subfloat[\textit{ETH3D}]{\includegraphics[height=8cm]{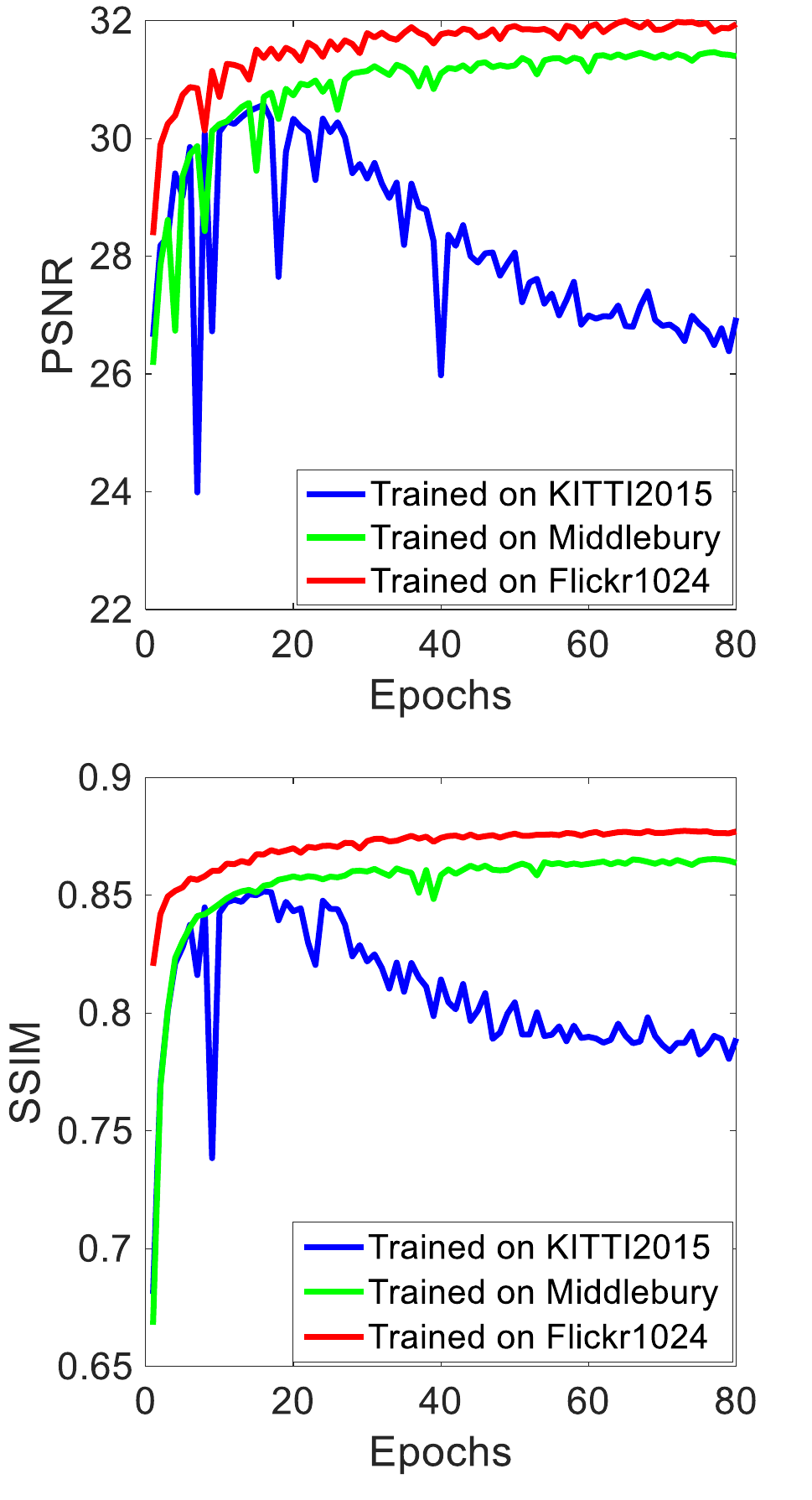}}

\caption{PSNR and SSIM values achieved by \textit{PASSRnet}\cite{wang2019learning} with different numbers of training epochs for 4$\times$~SR. Note that, the performance is evaluated on the test sets of (a) \textit{KITTI2015}, (b) \textit{Middlebury}, (c) \textit{Flickr1024}, and (d) \textit{ETH3D}, respectively. }\label{fig4}
\vspace{0.3cm}
\end{figure*}

 \subsection{Implementation Details}

 We use two state-of-the-art stereo SR methods (i.e., \textit{StereoSR}\cite{jeon2018enhancing} and \textit{PASSRnet}\cite{wang2019learning}) in this experiment. These two methods are first trained on the \textit{KITTI2015}, \textit{Middlebury}, and \textit{Flickr1024} datasets, and then tested on the above three datasets and the \textit{ETH3D} dataset. For simplification, only 4$\times$ SR models are investigated. That is, stereo image pairs are first down-sampled by a factor of 4, and then super-resolved to their original resolutions. We compare the reconstructed image with the original image, and use PSNR and SSIM for performance evaluation.

 We used the codes of \textit{StereoSR}\cite{jeon2018enhancing} and \textit{PASSRnet}\cite{wang2019learning} released by their authors. Since the \textit{StereoSR} model trained on the \textit{Middlebury} dataset is available, we directly use this model in our experiment. For the other 5 unavailable models, we retrain these two SR methods.

\subsection{Results}

Tables~\ref{tab2} and \ref{tab3} present the results of \textit{StereoSR} and \textit{PASSRnet} trained with fixed training epochs. We can observe that both algorithms trained on the \textit{Flickr1024} dataset achieve the highest PSNR and SSIM values on all of the test sets as compared to those trained on the \textit{KITTI2015} and \textit{Middlebury} datasets. Specifically, the \textit{Flickr1024} dataset outperforms the second best datasets (\textit{KITTI2015} in Table~\ref{tab2} and \textit{Middlebury} in Table~\ref{tab3}) by 1.04 and 0.58 in average PSNR values, respectively. These results demonstrate that the \textit{Flickr1024} dataset can help to significantly improve the performance of stereo SR algorithms.

Moreover, we train \textit{PASSRnet}\cite{wang2019learning} with different training epochs, and further investigate the variation of PSNR and SSIM values. The results are shown in Fig.~\ref{fig4}, where each sub-figure illustrates the performance tested on a specific dataset. We can observe that the algorithm trained on the \textit{Flickr1024} dataset achieves the highest PSNR and SSIM values with any number of training epochs. For the models trained on the \textit{KITTI2015} dataset, their PSNR and SSIM curves suffer a downward trend. In contrast, the models trained on the \textit{Flickr1024} dataset can achieve a gradually improved performance with an increasing number of training epochs. These results demonstrate that, by using our dataset, a reasonable convergence can be steadily achieved and the over-fitting issue can be well addressed.

\section{Conclusion}\label{sec5}

In this paper, we introduce \textit{Flickr1024}, a large-scale dataset for stereo SR. The \textit{Flickr1024} dataset consists 1024 high-quality images and covers diverse scenarios. Both statistical comparisons and cross-dataset experiments demonstrate the effectiveness of our dataset. That is, the \textit{Flickr1024} dataset can be used to improve the performance of learning-based stereo SR methods. The \textit{Flickr1024} dataset can also help to boost the reseach in stereo super-resolution.

\section{Acknowledgment}\label{Acknowledgement}
The authors would like to thank \textit{Sascha Becher} and \textit{Tom Bentz} for the approval of using their cross-eye stereo photographs on \textit{Flickr}.

{\small
\bibliographystyle{ieee}
\bibliography{manuscript}
}

\end{document}